\documentclass[runningheads]{llncs}
\pdfoutput=1
\usepackage{cite}
\usepackage{amsmath,amssymb,amsfonts}
\usepackage{algorithmic}
\usepackage{graphicx}
\usepackage{textcomp}
\usepackage{xcolor}
\usepackage{diagbox}
\usepackage{makecell}
\usepackage{booktabs}
\usepackage[ruled,vlined]{algorithm2e}
\usepackage{caption}
\usepackage{siunitx}
\usepackage{wrapfig}
\begin{document}
\title{Human-Understandable Decision Making for Visual Recognition
}
%
%
\author{Xiaowei Zhou\inst{1,3} \and
Jie Yin\inst{2} \and 
Ivor Tsang\inst{1} \and
Chen Wang \inst{3} }
\authorrunning{X. Zhou et al.}
%
\institute{Australian Artificial Intelligence Institute, FEIT, University of Technology Sydney, Sydney, Australia \\
\email{Xiaowei.Zhou@student.uts.edu.au, Ivor.Tsang@uts.edu.au}
\and
Discipline of Business Analytics, The University of Sydney, Sydney, Australia
\email{jie.yin@sydney.edu.au}\\
 \and
Data61, CSIRO, Sydney, Australia\\
\email{Chen.Wang@data61.csiro.au}}
\maketitle              
\begin{abstract}
The widespread use of deep neural networks has achieved substantial success in many tasks. However, there still exists a huge gap between the operating mechanism of deep learning models and human-understandable decision making, so that humans cannot fully trust the predictions made by these models. To date, little work has been done on how to align the behaviors of deep learning models with human perception in order to train a human-understandable model. To fill this gap, we propose a new framework to train a deep neural network by incorporating the prior of human perception into the model learning process. Our proposed model mimics the process of perceiving conceptual parts from images and assessing their relative contributions towards the final recognition. The effectiveness of our proposed model is evaluated on two classical visual recognition tasks. The experimental results and analysis confirm our model is able to provide interpretable explanations for its predictions, but also maintain competitive recognition accuracy.

\end{abstract}
\section{Introduction}
Deep neural networks (DNNs) have made remarkable success in many areas such as computer vision, 
speech recognition, and natural language processing. 
Although DNNs have achieved human-level performance or even beaten humans on some tasks like object recognition or video games, their underlying operation mechanism still remains a ``black box" for humans to fully comprehend. As a result, humans cannot fully trust the predictions made by DNNs, particularly in life-critical applications, such as auto-driving, and medical diagnosis. Therefore, not only the academia, but also the industry is regarding interpretability as one of the most important components and even a must-have one for responsible use of deep learning models~\cite{ghorbani2019towards}. 

Interpretability of machine learning refers to the ability to explain or to present the results in understandable terms to a human~\cite{doshi2017towards}. There have been considerate research efforts on studying the interpretability of deep learning models. A stream of research has adopted a post-hoc approach, which yields explanations by visualizing important features~\cite{adebayo2018sanity,shrikumar2017learning,sundararajan2017axiomatic}, or by creating a surrogate model with high fidelity to the original model~\cite{ribeiro2016should,wu2018sharing}. However, feature visualisation based methods are found to be unreliable, as small perturbations to the input data can lead to dramatically different explanations~\cite{ghorbani2019interpretation,adebayo2018sanity}. Methods based on surrogate models do not reveal the essential process of the original model to enhance human understanding~\cite{laugel2019dangers}.

Recently, researchers have attempted to train an inherently interpretable model directly~\cite{hendricks2016generating,dong2017improving,zhang2018interpretable}. The aim is to learn a deep learning model from scratch that is able to give explanations. These methods often use auxiliary information as additional supervision to train the model. Such information is encoded in the loss function to regularize model training, aiming to learn certain mappings between latent features and human-understandable semantic information. Model prediction results are then augmented with auxiliary outputs, such as topic words~\cite{dong2017improving}, sentences~\cite{hendricks2016generating}, or object parts~\cite{dong2017improving}, to improve human understanding. However, due to the lack of connection between model learning and human-understandable decision making, how these models make decisions is still beyond the direct comprehension of humans. Even worse, there is no clue to assert whether or not decisions made by these models are reliable.

\begin{wrapfigure}{r}{0.5\textwidth}
\vspace{-0.5in}
\begin{center}
\centerline{\includegraphics[width=0.5\columnwidth]{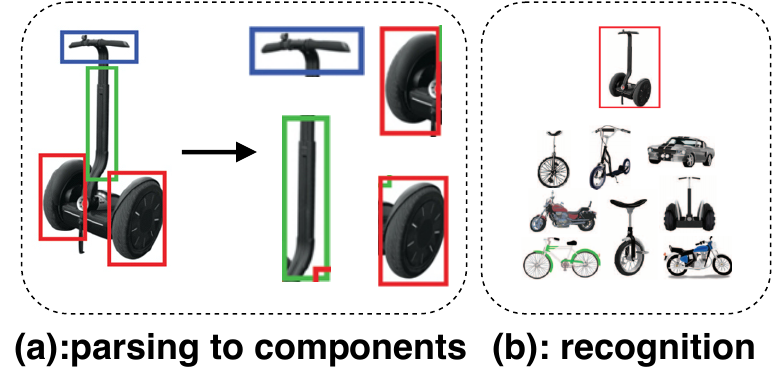}}
\caption{(a) Humans parse the object into different semantic components; (b) Results from different semantic components are aggregated for final recognition. This figure is adapted from \cite{lake2017building}.}
\label{fig:moti} 
\end{center}
\vspace{-0.5in}
\end{wrapfigure}

To fill this research gap, in this paper, we propose a new learning framework that makes predictions in a human-understandable way. Our core idea is inspired by key findings from cognitive science~\cite{biederman1987recognition,lake2017building} that humans make decisions based on their perception. As illustrated by Fig.~\ref{fig:moti}, humans parse an object into different components and then aggregate the results from these components to derive the final recognition. We call these semantic components as \textit{conceptual parts} in this paper. A question naturally comes into our mind: Can we build a deep learning model that makes predictions following this human-understandable way? If we can link model learning with this human decision making process, it is likely that the learned deep learning model can be better understood by humans.

To achieve this goal, we design a new learning framework in analogy to the process of human-understandable decision making. Our framework comprises an automatic concept partition model and a concept-based recognition model via mixture of experts~\cite{jacobs1991adaptive}. The concept partition model first parses the input images into meaningful conceptual parts associated with different semantics, such as different body parts of a bird or  sense organs of human faces. The features of each conceptual part are then passed to a respective expert model for recognizing the input image. A gate network is used to aggregate the predictions from different experts to obtain the final prediction results as well as the importance weights of different conceptual parts. The importance weights reflect the contributions of different conceptual parts to the final prediction. As a result, our designed model incorporates how humans perceive from images and is able to give human-understandable explanations. We show that our proposed model is able to achieve comparable or even better accuracy on two visual recognition tasks. More importantly, our model can provide human-understandable explanations by conceptual parts and their importance weights, thereby enhancing human understanding of how the model makes predictions. 


The contribution of this paper is three-fold:
\begin{itemize}
    \item We propose to inject human perception from images into model design and learning of DNN models to enable human-understandable decision making.
    \item Our new framework is able to give explanations through conceptual parts and their relative importance for the recognition result of each input image.
    \item Experiments on two visual recognition tasks verify the consistency between the learned explanations and human cognition.
\end{itemize}

\begin{figure*}[tbp]
\begin{center}
\centerline{\includegraphics[width=0.9\textwidth]{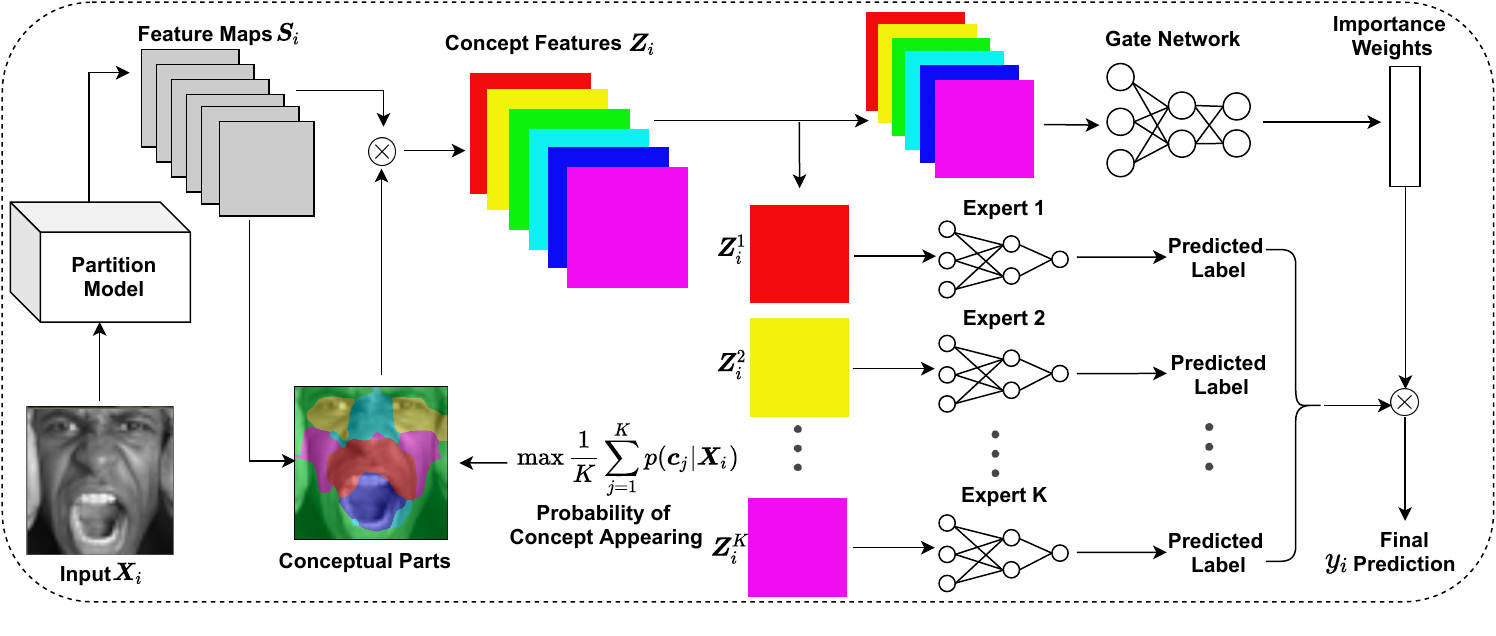}}
\vskip -0.1in
\caption{The overview of the proposed method. Our model learns to partition the input image into conceptual parts, whose features are passed to the expert networks to make predictions. The gate network learns the importance weights of different conceptual parts for the final prediction.}
\label{fig:structure} 
\end{center}
\vskip -0.35in
\end{figure*}

\section{Model Design with Human Perception}
Our proposed framework trains a deep learning model by injecting the prior of how humans perceive from images. The overview of the proposed framework is shown in Fig.~\ref{fig:structure}. Our model first learns to parse the input images into conceptual parts with different semantics. Features of conceptual parts are then fed to the concept-based recognition model to make the final prediction. 

\subsection{Concept Partition Model}

We train a concept partition model to automatically segment the input images $\mathbb{X}=\{\boldsymbol{X}_0, \cdots, \boldsymbol{X}_N\}$ into a set of conceptual parts with different semantics using only image-level labels. Following~\cite{huang2020interpretable}, we transform the concept partition problem into the problem of estimating the probability $p(\boldsymbol{c}_j|\boldsymbol{X}_i)$ that each concept $\boldsymbol{c}_j$ occurrs in the input image $\boldsymbol{X}_i$. The classical convolution neural networks, i.e., ResNet, are used as the backbone to build our partition model. We use feature maps $\boldsymbol{S}_i \in R^{D \times H \times W}$ from one layer of the neural network to estimate the probability whether or not each concept appears in an input image $\boldsymbol{X}_i$. Based on the feature maps $\boldsymbol{s}_{hw} \in R^{D}$ at position $(h,w)$ of $\boldsymbol{S}_i$, the probability $p_{h,w}^j$ of the $j$-th concept $\boldsymbol{c}_j  \in R^D$ occurring at the position $(h,w)$ is formulated as: 
\begin{equation}
p_{h,w}^{j}=\frac{\exp \left(-\left\|\left(\boldsymbol{s}_{hw}-\boldsymbol{c}_{j}\right) / \alpha_{j}\right\|_{2}^{2} / 2\right)}{\sum_{j} \exp \left(-\left\|\left(\boldsymbol{s}_{hw}-\boldsymbol{c}_{j}\right) / \alpha_{j}\right\|_{2}^{2} / 2\right)},
\end{equation}
where $\alpha_j \in (0,1)$ is a learnable smoothing factor for each concept $\boldsymbol{c}_j$; $\boldsymbol{c}_j$ is a vector representing the $j$-th concept, which can be considered as the center of a cluster; $p_{h,w}^{j}>0$ and $\sum_{j=1}^K p_{h,w}^{j}=1$; $K$ is the number of concepts.

Then, we can obtain the concept occurrence map $\boldsymbol{O} = [p_{h,w}^{j}] \in R^{K \times H \times W}$ by assembling all the $p_{h,w}^{j}$'s, which indicates the probability of each concept occurring at each position. At each position, we rank the occurrence probability of each concept and assign the concept with the highest probability as the one occurring at that position, i.e., $j^*=\arg\max _j \boldsymbol{O}$, where $j = 1, \cdots, K$ represents the index of concepts. Using this approach, we obtain the concept partition result for each input image. One example is given in Fig.~\ref{fig:structure}, where the image is parsed into different conceptual parts represented by different colors. 

As we have only image-level labels as the supervision to train the concept partition model, we obtain the concept features $\boldsymbol{Z}_i$ to predict image-level labels, based on the concept occurrence map $\boldsymbol{O}$ and feature maps $\boldsymbol{S}_i$. Let $\boldsymbol{Z}_i^j$ be the $j$-th dimension of $\boldsymbol{Z}_i$, representing concept features of the $j$-th concept, we have $\boldsymbol{Z}_{i}^j=\frac{\boldsymbol{t}_{i}^{j}}{\left\|\boldsymbol{t}_{i}^{j}\right\|_{2}},$ where $\boldsymbol{t}_{i}^{j}=\frac{1}{\sum_{hw} p_{hw}^{j}} \sum_{hw} p_{hw}^{j}\left(\boldsymbol{s}_{hw}-\boldsymbol{c}_{j}\right) / \sigma_{j}$. The concept features $\boldsymbol{Z}_i$ are passed to a classifier $h$ constructed with several convolutional layers and fully connected layers. The cross-entropy loss is used to train the concept partition model and the classifier. The classification loss is formulated as $l_{cls} = - \frac{1}{n} \sum_{i=1}^N \hat{y}_i \log (h(\boldsymbol{Z}_i))$, where $N$ is the number of input images, and $\hat{y}_i$ is the ground-truth label of image $\boldsymbol{X}_i$.

However, our exploration shows that using only the classification loss is insufficient to obtain meaningful concept partition results. Given the prior knowledge that all relevant concepts could occur in each image; for example, different body parts are likely to occur in most of bird images, we introduce a regularizer to incorporate such prior knowledge. Our goal is to maximize the probability of each concept appearing in the input images, which is formulated as:
\begin{equation}
   l_r = \min  \frac{1}{K*N}\sum_{i=1}^N \sum_{j=1}^K |\log (p(\boldsymbol{c}_j|\boldsymbol{X}_i) + \delta)|,
   \label{eq:loss_p}
\end{equation}
where $\delta$ is a small value (\num{1e-5}) for stable training; $p(\boldsymbol{c}_j|\boldsymbol{X}_i)$ is the probability of concept $\boldsymbol{c}_j$ occurring in the input image $\boldsymbol{X}_i$. It can be obtained through aggregating the probability of concept $\boldsymbol{c}_j$ occurring in each position, i.e.,  $p(\boldsymbol{c}_j|\boldsymbol{X}_i) = \max_{hw} \mathcal{G} \times \boldsymbol{O}_j$, where $\mathcal{G}$ is a 2D Gaussian kernel for smoothing, $\boldsymbol{O}_j \in R^{H \times W}$, $\max$ indicates the max pooling operation. Finally, the cross-entropy loss $l_{cls}$ and the loss $l_r$ in Eq.~(\ref{eq:loss_p}) are combined to train the concept partition model.

\subsection{Concept-based Recognition Model}

Based on concept features obtained from the concept partition model, we train a concept-based recognition model to perform the final prediction. Our concept-based recognition model involves a set of experts, each of which takes  features of each concept as input to make a prediction. These concept features represent different semantic concepts, such as mouth, eyes, and cheek in the facial expression recognition task. A gate network is used to aggregate the prediction results from different experts to produce the final recognition and to estimate the importance weights of each concept. The importance weights reflect different contributions made by distinct conceptual parts to the final recognition.

For a specific recognition task, suppose the input image $\boldsymbol{X}_i$ is transformed into concept features $\boldsymbol{Z}_i$ via the concept partition model. Each concept feature $\boldsymbol{Z}_i^j$ is recognized by an expert $f_j(\boldsymbol{Z}_i^j)$. The predictions by all experts are aggregated by the importance weights to get the final recognition. The importance weights of different experts are learned by a gate network $g(\boldsymbol{Z}_i)$. Formally, we build a recognition model that can be formulated as follows:
\begin{equation}
    \begin{aligned}
        f(\boldsymbol{x}_i) &= {
        \left[ \begin{array}{cccc}
        \boldsymbol{w}^1_i, & \boldsymbol{w}_i^j & \cdots & \boldsymbol{w}_i^K
        \end{array} 
        \right ]} \times 
        {\left [ \begin{array} {c} 
        f_1(\boldsymbol{Z}_i^1)\\
        f_j(\boldsymbol{Z}_i^j)\\
        \vdots \\
        f_K(\boldsymbol{Z}_i^K)
        \end{array}
        \right ]}. 
    \end{aligned}
    \label{eq:skeleton}
\end{equation}
Above, $f(\boldsymbol{x}_i)$ is the final predicted label for the input $\boldsymbol{X}_i$. The gate network $g(\boldsymbol{Z}_i)$ is parameterized by $[\boldsymbol{w}^1_i,  \boldsymbol{w}_i^j, \cdots, \boldsymbol{w}_i^K]$, where $\boldsymbol{w}_i^j$ is the weight of concept feature  
$\boldsymbol{Z}_i^j$ 
for the $j$-th concept in $\boldsymbol{X}_i$. 
$K$ is the number of experts, which is equal to the number of conceptual parts. 

Each expert $f_j(\boldsymbol{Z}_i^j)$ is constructed with the same network structure but is given different concept features as input. The gate network $g(\boldsymbol{Z}_i)$ is also a neural network, which takes all concept features $\boldsymbol{Z}_i$ as input. The last layer of the gate network is a softmax layer that produces the weights summed to one. The weight $\boldsymbol{w}_i^j$ can be regarded as importance weight of each concept for the final prediction. The learned importance weights facilitate humans to better understand how much contribution each conceptual part makes towards the final recognition.

To train the concept-based recognition model, we use the cross-entropy loss with a regularizer imposed on importance weights. Specifically, we use image-level labels as supervision to train each expert. The overall loss function for all experts is: $l_{ept} = - \frac{1}{n} \sum_{i=1}^N \sum _{j=1}^K \hat{y}_i \log (f_j(\boldsymbol{Z}_i^j))$, where $N$ is the number of input images, $\hat{y}_i$ is the ground true label of image $\boldsymbol{X}_i$ and $f_j(\boldsymbol{Z}_i^j)$ is the predicted label for image $\boldsymbol{X}_i$ by expert $j$. We also use the cross-entropy loss to train the gate network. Additionally, a constraint is imposed on the value of importance weight $\boldsymbol{w}_i$, preventing one weight from dominating the prediction and generating meaningless importance weights. $\gamma$ is the weighting factor for balancing the two terms. The overall loss function for training the gate network is formulated as:
\begin{equation}
    \begin{aligned}
        l_{g} = - \frac{1}{N} \sum_{i=1}^N \hat{y}_i \log \left (f(\boldsymbol{Z}_i) \right )
        + \gamma \frac{1}{K*N}\sum_{i=1}^N \sum_{j=1}^K \Bigg\|\boldsymbol{w}_i^j - \frac{1}{K}\Bigg\|_2^2.
    \end{aligned}
    \label{eq:loss2}
\end{equation}

\section{Experimental Evaluation}
To validate the effectiveness of our proposed model, we perform two visual recognition tasks with varying task difficulty. We use ResNet101 as the backbone to parse the images into conceptual parts. Concept features of each conceptual part are further used to train each expert and the gate network. 


\subsection{Facial Expression Recognition} 
We begin with the facial expression recognition task on the FER-2013 dataset~\cite{goodfellow2013challenges}.
It consists of 28,709 training images, 3,589 public test images, and 3,589 private test images. The original face images in the dataset are grey-scale images of size $48\times48$. All images are categorized into 7 classes. 
Ian~\cite{goodfellow2013challenges} reported that human accuracy on FER-2013 was around 65\%.

\begin{wraptable}{r}{0.48\textwidth}
    \vspace{-0.8cm}
    \centering
    \tabcolsep 12pt
    \captionof{table}{Classification accuracy on FER2013 test dataset.}
    \label{tab:claFER}
    \begin{tabular}{lr}
    \toprule
    Model & Accuracy (\%) \\ 
    \midrule
    ResNet101    & 71.44 \\ 
    ResNet50 & 70.47 \\ 
    ResNet18    &  69.74 \\ 
    Vgg16\_bn    & 70.35 \\ 
    InceptionV3 &  68.99 \\ 
    \midrule
    Our model   &   \textbf{73.67} \\ 
    \bottomrule
    \end{tabular}
    \vspace{-0.8cm}
\end{wraptable}

\vspace{-3mm}
\subsubsection{Classification Performance.} 
We train our model using training images resized into $224\times224$ with three channels. We set the learning rate $\lambda$ as 1e-4; use the SGD optimizer with momentum as 0.9 and weight decay as 5e-4; set weighting factor $\gamma$ as 1.0; set number of experts and number of conceptual parts as 6; set the maximum iteration as 200. We compare our model with several classical baseline models (i.e. ResNet, VGG, Inception V3) on the FER2013 private test dataset. The classification results are summarized in Table~\ref{tab:claFER}. All classification results are obtained by training from pre-trained weights on ImageNet. From this table, we can see that our model achieves the best classification accuracy. This validates the effectiveness of our model in facial expression recognition, beating the human accuracy of $65$\% on this dataset.

\vspace{-3mm}
\subsubsection{Explanation Results.}
\begin{figure}[tbp]
\begin{center}
\centerline{\includegraphics[width=0.92\textwidth]{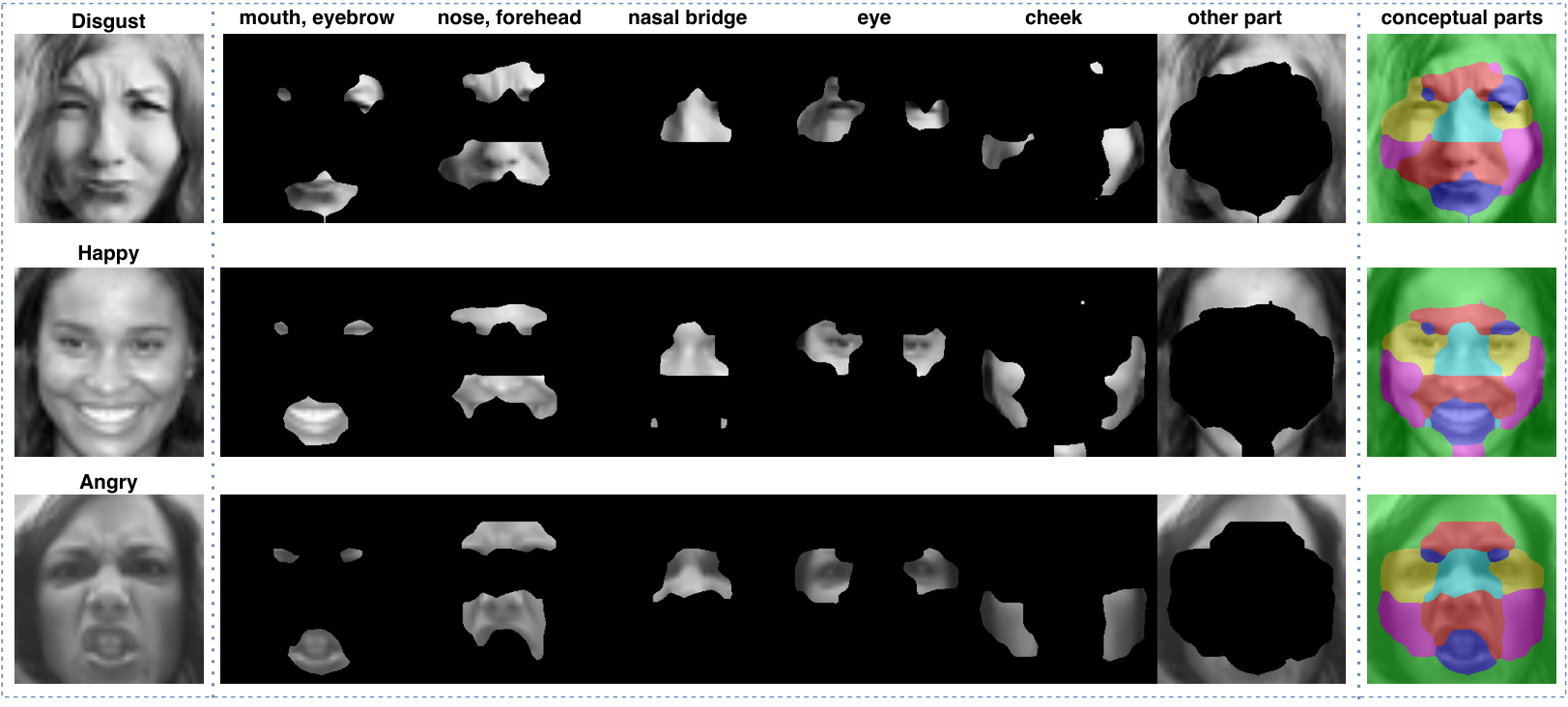}}
\vspace{-0.1in}
\caption{The conceptual parts learned by our model on FER2013 dataset.}
\label{fig:FER}
\end{center}
\vskip -0.3in
\end{figure}

After our model is trained, the prediction results can be explained through partitioned conceptual parts and their importance weights. Fig.~\ref{fig:FER} shows three example images from FER2013 and the learned conceptual parts. The first-column images are the original images with their class labels on the top. The columns 2-7 are the identified conceptual parts with the concept name on the top. The images in the last column are the conceptual parts collated with each color indicating one concept. When we set the number of conceptual parts as 6, the partition model parses each image into 6 parts with different semantics, i.e., nose/forehead, eye, nasal bridge, mouth/eyebrow, cheek, and other part. These conceptual parts are easy for humans to recognize and understand. Furthermore, the partition results are consistent for different input images. That is to say, the concept partition model is able to identify meaningful concepts for the recognition task. 

\begin{wrapfigure}{r}{0.49\textwidth}
    \vspace{-1.4cm}
    \centering
    \includegraphics[width=0.48\textwidth]{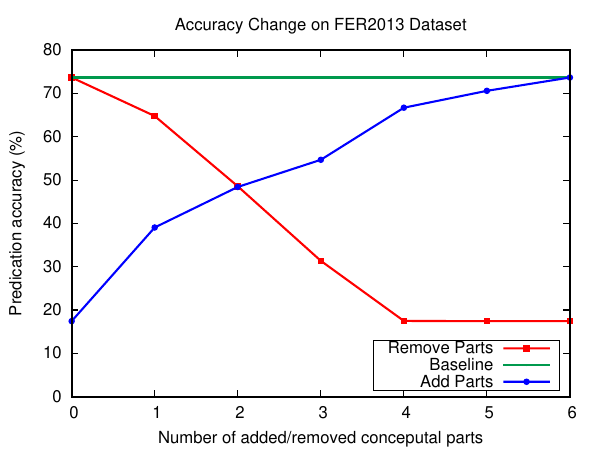}
    \vspace{-0.2cm}
    \captionof{figure}{Classification accuracy vs. adding/removing important conceptual parts.}
    \label{fig:acc_FER}
    \vspace{-0.8cm}
\end{wrapfigure}

To show the importance of different conceptual parts for recognition, we calculate the average weight of each conceptual part for test images on FER-2013. Table~\ref{tab:gammaExp} shows the average weight of each conceptual part. When $\gamma=1.0$, conceptual parts are ranked as mouth/eyebrow, nose/forehead, nasal bridge, eye, cheek and other parts, according to their importance weights.

\begin{table}[tbp]
\centering
\caption{Average weight of conceptual parts learned on FER2013.}
\label{tab:gammaExp}
\begin{tabular}{p{0.15\linewidth}|c|c|c|c|c|c}
\toprule
Conceptual Parts & mouth/eyebrow & nose/forehead & nasal bridge & eye & cheek & other parts \\ 
\midrule
Avg. Weight & 0.2127 & 0.1912 & 0.1856 & 0.1539 & 0.1444 & 0.1120 \\
\bottomrule
\end{tabular}
\vskip -0.1in
\end{table}

We further test the effect of adding or removing the learned conceptual parts according to their associated rank in Table~\ref{tab:gammaExp}. As shown in Fig.~\ref{fig:acc_FER}, the green line indicates the baseline classification accuracy (73.67\%) of our model with all concept features as input. The red line (with squares) shows the accuracy loss caused by gradually removing the most important conceptual parts from the input. As can be seen, the accuracy drops significantly to 17.47\%, after the top 4 important conceptual parts are removed. The blue curve (with dots) shows the changes in accuracy when important conceptual parts are added. As we gradually add the most important conceptual parts, the prediction accuracy increases from 17.44\% to 73.67\%. With only 4 important conceptual parts, our method achieves the accuracy of 66.70\%. This validates the effectiveness of the learned importance weights and their contribution towards the overall classification performance.

Lastly, we also conduct human evaluation to verify the consistency between our model and human decision making. We asked 30 participates to answer two questions: Q1, how important are the learned conceptual parts? Q2, what is the importance order of conceptual parts for them to make decisions? We randomly selected 60 images from FER2013 test dataset to conduct human evaluation with results summarized as follows. Firstly, we calculate the average importance score in Q1 across different images and participants. Out of 5 points, we obtain 4.21 points on average. That means that participants think the predictions made by our trained model is consistent with their perception for visual recognition. Secondly, we calculate the recall of top 4 conceptual parts considered important by participants and also selected as the top 4 important ones by our model. The average recall is $78.17\%$. This also exhibits a high level of consistency between our model predictions and human judgements.


\subsection{Fine-grained Bird Classification}

Next, we conduct experiments on a more difficult, fine-grained bird classification task. CUB\_200\_2011~\cite{WahCUB_200_2011} is a fine-grained bird dataset that contains 11,788 images of 200 bird species. The training dataset has 5,994 images, and the rest of 5,774 images are for testing.

\begin{wraptable}{r}{0.48\textwidth}
    \vspace{-0.9cm}
    \centering
    \tabcolsep 6pt
    \captionof{table}{Classification accuracy on CUB\_200\_2011 test dataset.}
    \label{tab:claBird}
    \begin{tabular}{lr}
    \toprule
    Model & Accuracy (\%) \\ 
    \midrule
    ResNet101    & 85.02 \\ 
    ResNet50 &  84.45 \\ 
    ResNet18    &  81.67 \\ 
    Vgg16\_bn    &  83.81 \\ 
    InceptionV3 &  84.05 \\ 
    MC (TIP2020)~\cite{chang2020devil} & \textbf{87.30} \\
    \midrule
    Our model  &  \textbf{86.66} \\  
    \bottomrule
    \end{tabular}
    \vspace{-0.5cm}
\end{wraptable}

\vspace{-3mm}
\subsubsection{Classification Performance.}
We train our model on CUB\_200\_2011 training dataset, where images are resized into $448\times448$ as input. We set the learning rate $\lambda$ as 5e-5; set number of experts and number of conceptual parts as 5; other parameters are the same as on FER-2013. We compare classification performance of our model with ResNet, VGG, InceptionV3, and MC Loss method~\cite{chang2020devil} on the test dataset, as reported in Table~\ref{tab:claBird}. All results are obtained by training from pre-trained weights on ImageNet. We can see that our model performs better than the other three classical models except for MC method that is specially designed for fine-grained classification. Overall, our model is able to achieve better accuracy than classical models and comparable accuracy with state-of-the-art model.

\begin{figure}[tbp]
\begin{center}
\centerline{\includegraphics[width=\linewidth]{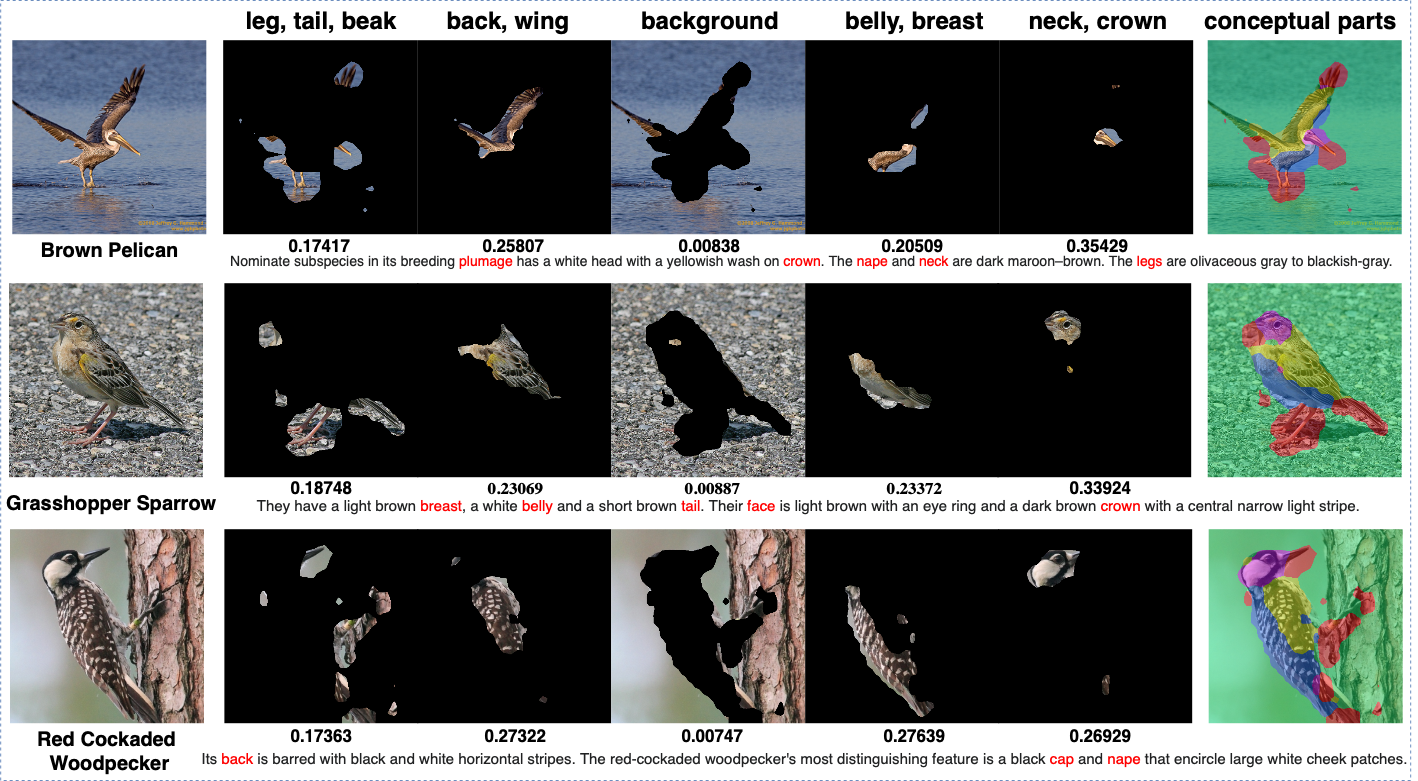}}
\caption{The conceptual parts learned by our model on the CUB\_200\_2011 dataset. The text description underneath conceptual concepts are the definitions of the corresponding bird species from Wikipedia.}
\label{fig:BIRD}
\end{center}
\vskip -0.4in
\end{figure}

\vspace{-3mm}
\subsubsection{Explanation Results} 

Fig.~\ref{fig:BIRD} shows several example images from CUB\_200\_2011 and their important conceptual parts learned for classification. The images in the first column are the original images with their class labels on the bottom. The last column shows conceptual parts identified with each color indicating one concept. The columns 2-6 are conceptual parts with the concept definition on the top, where the values under the partition results are the importance weights of different concepts for classifying that image. The description texts are the definitions of the corresponding bird species from Wikipedia, where key features of different bird species are highlighted in red. 

We notice that the important conceptual parts learned by our model have a high degree of consistency with key features of bird species given by human experts. For example, for the first image in the second row in Fig.~\ref{fig:BIRD}, labeled as \textit{Grasshopper Sparrow}, crown, face, breast, belly, and tail are the discriminative features that human experts use to define this bird species. These corresponding conceptual parts are also identified and attributed with high weights by our model; crown/neck and belly/breast are ranked by our method as the first and second most important concepts for classification. 

Table~\ref{tab:ave_w_bird} lists the average importance weights of conceptual parts learned on CUB\_200\_2011. We find that neck/crown is the most important concept for recognition. In contrast, background is the least important concept. We also observe that the importance weight of concept leg/tail/beak is nearly 0, when $\gamma$ is 0. This is inconsistent with the definition of bird species given by human experts (see Fig.~\ref{fig:BIRD}). This proves the necessity of adding the constraint in Eq.(\ref{eq:loss2}) that prevents certain weight from dominating the parameter estimation.

\begin{wrapfigure}{r}{0.49\textwidth}
    \vspace{-0.6cm}
    \centering
    \includegraphics[width=0.49\textwidth]{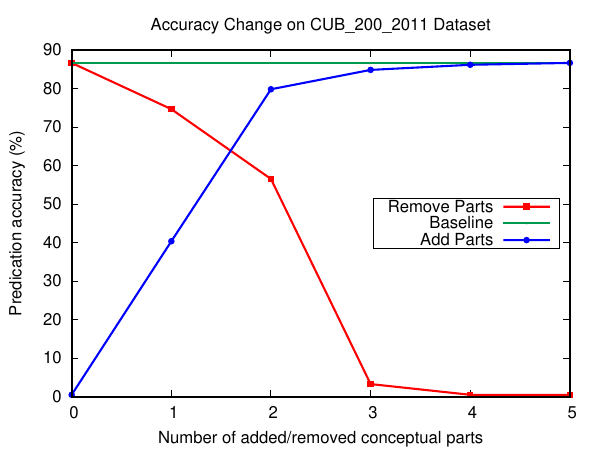}
    \vspace{-0.5cm}
    \captionof{figure}{Classification accuracy vs. adding/removing important conceptual parts.}
    \label{fig:acc_bird}
    \vspace{-0.7cm}
\end{wrapfigure}

Based on the importance ranking of conceptual parts ($\gamma=1$), we study the influence on recognition accuracy by adding or removing important conceptual parts. As shown in Fig.~\ref{fig:acc_bird}, the baseline accuracy (86.66\%) achieved by our model is plotted as the green line. Again, this is the result using all concept features as input. As indicated by the red line (with squares), when we remove the top 3 important conceptual parts, the prediction accuracy drops markedly from 86.66\% to 3.26\%. The prediction accuracy shown by the blue line (with dots) increases from 0.5\% to 79.84\%, when the top 2 important conceptual parts are added. All results validate the effectiveness of our learned importance weights in quantifying the contributions of different conceptual parts towards the final classification.

\begin{table}[tbp]
\centering
\caption{Average weight of conceptual parts learned on CUB\_200\_2011.}
\label{tab:ave_w_bird}
\begin{tabular}{c|c|c|c|c|c}
\toprule
Conceptual Parts & leg/tail/beak & back/wing & background & belly/breast & neck/crown \\ 
\midrule
$\gamma=$1 & 0.1760 & 0.2461 & 0.0079 & 0.2404 & 0.3296 \\
\midrule
$\gamma=$0 & 0.0044 & 0.2283 & 0.0022 & 0.2611 & 0.5039 \\
\bottomrule
\end{tabular}
\vskip -0.1in
\end{table}

\section{Discussions on Related Work}

Post-hoc explanation is the most popular method for interpreting deep learning models. This branch of methods can be divided into two categories: 1) \textit{important feature visualization}, which yields explanations by identifying and visualizing important features for input features. \cite{sundararajan2017axiomatic} proposed a method called integrated gradients to attribute the important features in input images. Grad-cam++~\cite{chattopadhay2018grad} used the gradients of each class as weights to combine the features of the last layer to get important features. However, these methods are found to be unreliable; small perturbations to the input data can lead to dramatically different explanations~\cite{ghorbani2019interpretation,adebayo2018sanity}. 2) \textit{surrogate models}, which create a surrogate model that can be more easily understood to mimic the performance of the original model. \cite{ribeiro2016should} tried to train a linear model to mimic the behaviors of the original deep learning models locally and used the learned model to explain the decisions of deep learning models. RICE~\cite{paccaci2019did} was proposed to give explanations for the target model by synthesizing logic program. However, methods based on surrogate models do not reveal how a decision is made by the original model~\cite{laugel2019dangers}.

Training an inherently interpretable model is another recent theme for interpreting deep learning models. \cite{hendricks2016generating} trained a deep classifier and an additional text generator to obtain classification results as well as text explanations for the original classifier. \cite{zhang2018interpretable} attributed the filters in high layers to different object parts by adding a mutual information loss between object parts and filters, when training a classifier. Semantic information was used as additional supervision in~\cite{dong2017improving} to train a video caption model, and the trained model could give the top important topics for explaining the results. These models augment their prediction results with auxiliary outputs, such as topic words~\cite{dong2017improving}, sentences~\cite{hendricks2016generating}, or object parts~\cite{dong2017improving}, to improve human understanding. However, due to disconnection between model learning and human decision making, it is still rather difficult for humans to fully comprehend how these models make predictions.

Our method is built upon semantic concepts for learning the partition and recognition model. We focus on training an interpretable model from a new angle of view; we inject human perception into the model design and learning process, yielding explanations that are more consistent with human cognition. Although there are many part-based object recognition methods~\cite{mordan2019end,zhang2016spda}, 
they solely aim at improving classification accuracy, without considering the interpretability of such models. In addition, the existing part-based methods either partition images into object-level parts rather than concept-level parts, or rely on pre-trained segmentation models with both image-level class labels and segment labels.

\section{Conclusion}
We proposed a new learning framework to train a deep learning model that makes predictions in a human-understandable way. Inspired by cognitive science, our framework is composed of a concept partition model, which learns conceptual parts with different semantics, and a concept-based recognition model, which makes the final prediction as well as yields relative importance weights of conceptual parts. Our method is able to provide human understandable explanations, because its design is more aligned with human-understandable decision making. Experiments on two visual recognition tasks showed that our method compares favourably to state-of-the-art methods on recognition accuracy, but also provides explanations that are highly consistent with human perception. For future work, we will investigate how to design our model in more complex classification tasks where the recognition by parts assumption may not hold.\\

\noindent \textbf{Acknowledgments.} This work is partially supported by ARC under Grant DP180100106 and DP200101328. Xiaowei Zhou is supported by a Data61 Student Scholarship from CSIRO.

\bibliographystyle{splncs04}
\bibliography{samplepaper.bbl}
%




\end{document}